\pdfoutput=1

\documentclass[11pt]{article}
\usepackage[]{naacl}
\usepackage{times}
\usepackage{latexsym}
\usepackage[T1]{fontenc}
\usepackage[utf8]{inputenc}
\usepackage{microtype}
\usepackage{helvet}

\usepackage{CJKutf8}
\usepackage{graphicx}
\usepackage{array}
\usepackage{arydshln}
\usepackage{multirow}
\newcommand{\minitab}[2][l]{\begin{tabular}{#1}#2\end{tabular}}
\usepackage{bm}
\usepackage{amssymb}
\usepackage{amsmath}
\usepackage{makecell}
\usepackage{wrapfig}
\usepackage{lipsum}  
\usepackage{etoolbox}
\usepackage{comment}
\usepackage{enumitem}

\usepackage{booktabs}
  
\usepackage{microtype}
\usepackage{csquotes}
\usepackage{bm}
\usepackage{amsmath,amssymb, amsfonts,mathtools}
\usepackage{soul}
\usepackage{adjustbox}
\usepackage[]{todonotes}
\usepackage[capitalize]{cleveref}
\usepackage{xcolor}
\usepackage{hyperref}
\usepackage{appendix}
\usepackage{svg}
\AtBeginEnvironment{appendices}{\crefalias{section}{appendix}}
\BeforeBeginEnvironment{appendices}{\clearpage}

\Crefname{appsec}{appendix}{appendices}

\usepackage{caption}
\usepackage{subcaption}
\usepackage{enumerate}
\usepackage{scalerel}
\usepackage[ruled,vlined]{algorithm2e}
\usepackage{linguex}
\usepackage{relsize}
\usepackage{array}
\usepackage{booktabs}
\usepackage{array}
\usepackage{makecell}
\usepackage{relsize}
\usepackage{multirow}

\usepackage[font=small,labelfont=bf]{caption}

\usepackage{cancel}
\usepackage{inconsolata}

\usepackage[utf8]{inputenc}
\definecolor{darkorange}{rgb}{1, 0.549, 0}
\BeforeBeginEnvironment{appendices}{\clearpage}

\usepackage{todonotes}
\newcommand{\note}[4][]{\todo[author=#2,color=#3,size=\scriptsize,fancyline,caption={},#1]{#4}} 

\newcommand{\eleanor}[2][]{\note[#1]{eleanor}{yellow}{#2}\xspace}

\newcommand{\Eleanor}[2][]{\eleanor[inline,#1]{#2}\noindentaftertodo}
\definecolor{babyblue}{rgb}{0.54, 0.81, 0.94}

\usepackage{stmaryrd}

\newcommand{\Verb}[1]{\textcolor{teal}{\textbf{\textit{#1}}}}
\newcommand{\Wrong}[1]{\underline{#1}}
\newcommand{\VerbWrong}[1]{\textcolor{teal}{\textbf{\textit{\underline{#1}}}}}
\newcommand{\Qiao}[1]{\textcolor{red}{\textbf{#1}}}
\newcommand{\Husband}[1]{\textcolor{blue}{\textbf{#1}}}
\newcommand{\Omit}[1]{$\llbracket$\textbf{#1}$\rrbracket$}
\newcommand{\orangeOmit}[1]{\textcolor{orange}{\Omit{#1}}}
\newcommand{\QiaoOmit}[1]{\textcolor{red}{\Omit{#1}}}

\newcommand{\gray}[1]{\textcolor{gray}{#1}}

\newcommand{\defn}[1]{\textbf{#1}}
\newcommand{\R}{\mathbb{R}}

\newcommand{\spans}{\mathsf{spans}}
\newcommand{\discourse}{\mathfrak{D}}
\newcommand{\discourseSet}{\mathbb{D}}

\newcommand{\vocab}{\mathcal{V}}
\newcommand{\Un}{S}

\newcommand{\NA}{\mathrm{NA}}

\newcommand{\category}{\mathbf{C}}
\newcommand{\categories}{\mathbf{C}}
\newcommand{\allcategories}{\mathbf{C}}

\newcommand{\categorizing}{\mathsf{cat}}
\newcommand{\similarity}{\bm{\mathsf{sim}}}
\newcommand{\countFun}{\bm{\mathsf{count}}}

\newcommand{\NoError}{\textsc{no error}}
\newcommand{\Document}{\textsc{document}}
\newcommand{\Sentence}{\textsc{sentence}}
\newcommand{\Inconsistency}{\textsc{inconsistency}}
\newcommand{\NamedEntity}{\textsc{named entity}}
\newcommand{\Entity}{\textsc{entity}}
\newcommand{\Tense}{\textsc{tense}}
\newcommand{\Ellipsis}{\textsc{ellipsis}}
\newcommand{\Pronoun}{\textsc{pronoun}}
\newcommand{\Others}{\textsc{other}}
\newcommand{\Ambiguity}{\textsc{ambiguity}}
\newcommand{\DiscourseMarker}{\textsc{dm}}

\newcommand{\entities}{\mathcal{E}}
\newcommand{\verbs}{\mathcal{V}}
\newcommand{\pronouns}{\mathcal{P}}
\newcommand{\dms}{\mathcal{M}}

\newcommand{\key}[1]{\texttt{#1}}
\newcommand{\Masculine}{\key{masculine}}
\newcommand{\Feminine}{\key{feminine}}
\newcommand{\Neuter}{\key{neuter}}
\newcommand{\Epicene}{\key{epicene}}

\usepackage{xspace}
\newcommand{\BWB}{\textsc{bwb}\xspace}

\newcommand{\SRC}{\textsc{src}}
\newcommand{\REF}{\textsc{ref}}
\newcommand{\MT}{\textsc{mt}}
\newcommand{\MTa}{\textsc{mta}}
\newcommand{\MTb}{\textsc{mtb}}
\newcommand{\SMT}{\textsc{smt}}
\newcommand{\OMTa}{\textsc{omt-a}}
\newcommand{\OMTb}{\textsc{omt-b}}
\newcommand{\OMTc}{\textsc{omt-c}}
\newcommand{\MTS}{\textsc{mt-s}}
\newcommand{\MTD}{\textsc{mt-d}}
\newcommand{\CADec}{\textsc{cadec}}
\newcommand{\CTX}{\textsc{ctx}}
\newcommand{\PE}{\textsc{pe}}

\newcommand{\BLEU}{\textsc{bleu}\xspace}
\newcommand{\BlonD}{\textsc{BlonDe}\xspace}
\newcommand{\dBlonD}{\textsc{BlonD-d}\xspace}
\newcommand{\BlonDp}{\textsc{BlonD+}\xspace}

\newcommand{\METEOR}{\textsc{meteor}}
\newcommand{\TER}{\textsc{ter}}
\newcommand{\ROUGEL}{\textsc{rouge-l}}
\newcommand{\CIDEr}{\textsc{CIDEr}}
\newcommand{\LC}{\textsc{lc}}
\newcommand{\RC}{\textsc{rc}}

\newcommand{\Skip}{\textsc{skip}}
\newcommand{\Aver}{\textsc{aver}}
\newcommand{\Vector}{\textsc{vector}}
\newcommand{\Greedy}{\textsc{greedy}}

\newcommand{\fluency}{\textsc{fluency}}
\newcommand{\adequacy}{\textsc{adequacy}}
\newcommand{\testset}[1]{\textsc{part#1}}
\newcommand{\rater}[1]{\textsc{rater#1}}
\newcommand{\diff}[2]{$\Delta$(#1, #2)}

\newcommand{\myFontSize}{\scriptsize}
\def\stripzero#1{\expandafter\stripzerohelp#1}
\def\stripzerohelp#1{\ifx 0#1\expandafter\stripzerohelp\else#1\fi}

\newcommand{\msra}{2}
\newcommand{\uzh}{3}
\newcommand{\ethz}{1}
\newcommand{\Langboat}{4}

\title{\BlonD: An Automatic Evaluation Metric for\\ Document-level Machine Translation}
\author{Yuchen Eleanor Jiang$^{\ethz}$\thanks{\hspace{1.5mm}Most of the work was done while the first author was an intern at Microsoft Research Asia.}~\;~Tianyu Liu$^{\ethz}$~\;~Shuming Ma$^{\msra}$~\;~Dongdong Zhang$^{\msra}$ ~\;~Jian Yang$^{\msra}$ \\
\textbf{Haoyang Huang$^{\msra}$~\;~Rico Sennrich$^{\uzh}$~\;~Mrinmaya Sachan$^{\ethz}$~\;~Ryan Cotterell$^{\ethz}$~\;~Ming Zhou$^{\Langboat}$}\\
  $^{\ethz}$ETH Z\"{u}rich~\;~$^{\msra}$Microsoft Research Asia~\;~$^{\uzh}$Universit{\"a}t Z\"{u}rich~\;~$^{\Langboat}$Langboat.com \\
  
  \texttt{\{\href{mailto:yucjiang@ethz.ch}{yucjiang},\href{mailto:tianyu.liu@inf.ethz.ch}{tianyu.liu},\href{mailto:ryan.cotterell@inf.ethz.ch}{ryan.cotterell},\href{mailto:mrinmaya.sachan@inf.ethz.ch}{mrinmaya.sachan}\}@inf.ethz.ch }\\
  \texttt{\{\href{mailto:shuming.ma@microsoft.com}{shuming.ma},\href{mailto:dongdong.zhang@microsoft.com}{dongdong.zhang},\href{mailto:t-jianya@microsoft.com}{t-jianya},\href{mailto:haohua@microsoft.com}{haohua}\}@microsoft.com}\\ 
  \texttt{\href{mailto:sennrich@cl.uzh.ch}{sennrich@cl.uzh.ch}}~\;~
  \texttt{\href{mailto:ming.zhou@chuangxin.com}{ming.zhou@chuangxin.com}}
}

\date{}
\begin{document}
\maketitle

\begin{abstract}
Standard automatic metrics, e.g., \BLEU{}, are not reliable for document-level MT evaluation. They can neither distinguish document-level improvements in translation quality from sentence-level ones, nor identify the discourse phenomena that cause context-agnostic translations.
This paper introduces a novel automatic metric \BlonD\footnote{\textbf{B}i\textbf{l}ingual Evaluati\textbf{on} of \textbf{D}ocum\textbf{e}nt Translation.} to widen the scope of automatic MT evaluation from the sentence to the document level.
\BlonD\ takes discourse coherence into consideration by categorizing discourse-related spans and calculating the similarity-based F1 measure of categorized spans. We conduct extensive comparisons on a newly constructed document-level translation dataset \BWB{}. 
The experimental results show that \BlonD{} possesses better selectivity and interpretability at the document-level, and is more sensitive to document-level nuances. 
In a large-scale human study, \BlonD\ also achieves significantly higher Pearson's $r$ correlation with human judgments compared to previous metrics.
\newline
\newline
\hspace{.5em}\includegraphics[width=1.25em,height=1.25em]{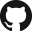}\hspace{.75em}\parbox{\dimexpr\linewidth-2\fboxsep-2\fboxrule}{\url{https://github.com/EleanorJiang/BlonDe}}

%

\end{abstract}

\section{Introduction}

Over the past years neural machine translation (NMT) models have become the models of choice in Machine Translation  \citep[MT;][\textit{inter alia}]{luong-etal-2015-effective,NIPS2017_7181,ctx}.
Although some recent work~\cite{hassan2018achieving, popel-2018-cuni, bojar-etal-2018-findings} suggests that NMT has achieved human parity at the sentence level, the reliability of these human-parity claims was quickly contested by \citet{laubli-etal-2018-machine, laubli2020set}, showing that there is a larger difference between human and machine translation quality when inter-sentential context is taken into account.

\begin{figure}[t]
    \centering
    \vspace{-10pt}
   \includegraphics[width=0.48\textwidth,trim={0.5cm 0cm 0.1cm 0.6cm} ,clip]{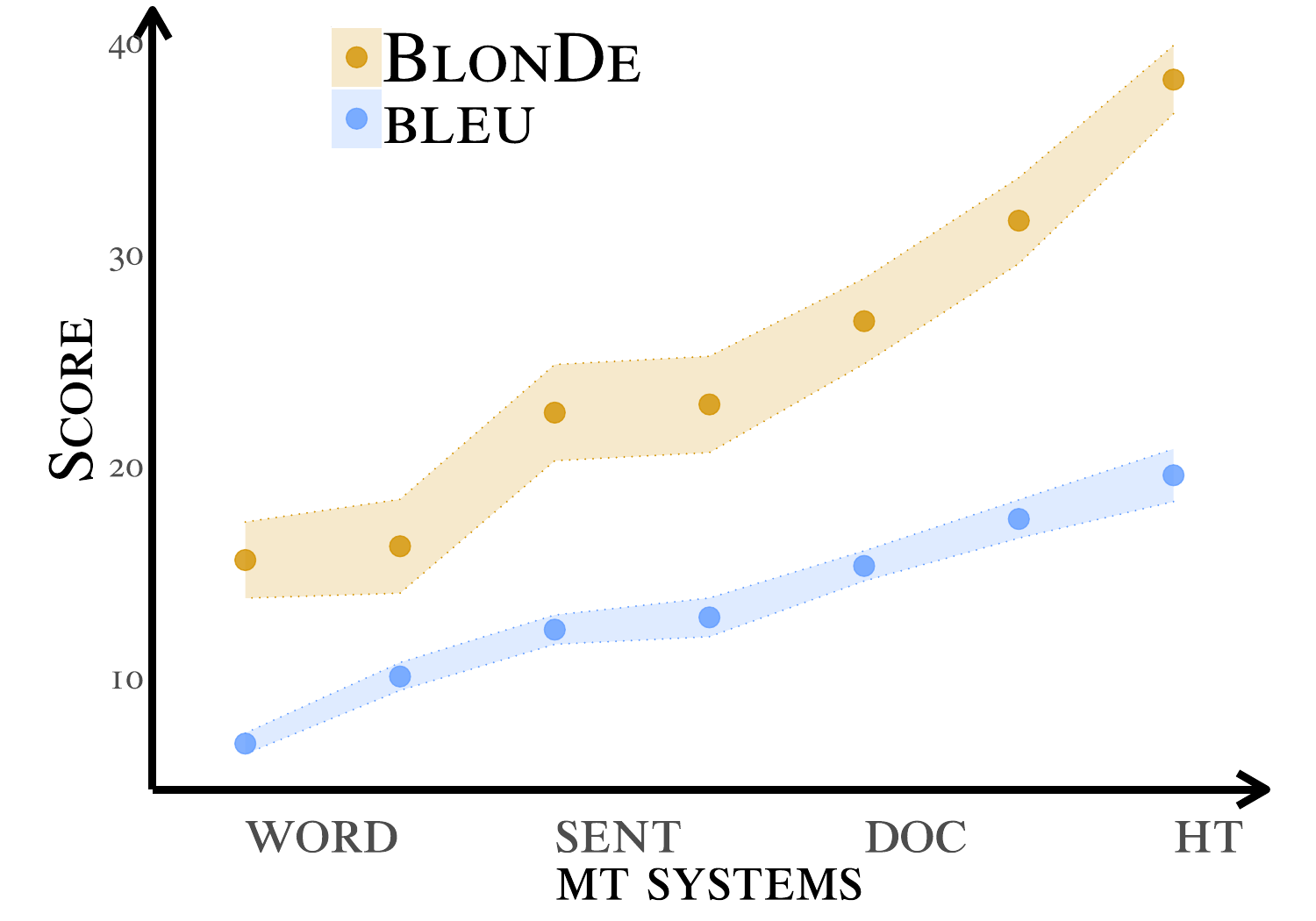}
    \vspace{-6pt}
    \caption{\BlonD{} is more selective than \BLEU{} for document-level MT, and also shows a larger quality difference between human and machine translations. \looseness=-1}
    \vspace{-10pt}
\end{figure}

Therefore, document-level machine translation has received increased attention in the MT community. 
However, despite various modeling advances, the MT community still lacks an efficient and effective evaluation metric for document-level translation.
Standard evaluation metrics for MT, e.g., \BLEU~\cite{BLEU}, \textsc{ter}~\cite{TER} and \METEOR~\cite{Meteor}, focus on the quality of translations at the sentence level and do not consider discourse-level features.

Thus, test suites that perform context-aware evaluation by targeting characteristic discourse-level phenomena have been proposed~\citep[][\textit{inter alia}]{DiscoMT, PROTEST, burchardt2017linguistic, isabelle-etal-2017-challenge, rios-gonzales-etal-2017-improving, muller-etal-2018-large,bawden-etal-2018-evaluating,voita-etal-2019-good, guillou-hardmeier-2018-automatic} for document-level MT. 
However, such test suites need to be re-created for new domains or even language pairs, and their construction can be very labor-intensive.
We still lack an easy-to-use automatic metric that can reliably discriminate the quality of document-level translation.\looseness-1


In this paper, we curate a large-scale document-level parallel corpus (\BWB) from heterogeneous data sources, and quantify document-level translation mistakes by performing a large human study. 
On this dataset, we found that inconsistency,\footnote{By inconsistency we mean the mistakes related to coreference and lexical cohesion~\cite{carpuat,guillou-2013-analysing}.} ellipsis, and ambiguity were the most noticeable phenomena critical for document-level MT, together amounting to 86.73\% of MT mistakes.
Based on this analysis, we propose \BlonD, an automatic metric that evaluates translation quality at the document level.
At the core of the metric is the similarity-based bijection between subsets of reference and system categories, e.g., pronouns, inflected forms, discourse relations and lexicons, and phrases, e.g., named entities. 
It computes recall, precision and F1, along with the corresponding measure of $n$-grams. Furthermore, \BlonD\ can incorporate human annotation easily by computing scores of human-annotated categories in the same way.\looseness=-1

We compare \BlonD\ with 11 other metrics and demonstrate that \BlonD\ is better at distinguishing between context-aware and context-agnostic MT systems.  
We also observe that the degree to which \BlonD{} correlates with sentence-level metrics (e.g., \BLEU) is lower than the degree to which the sentence-level metrics correlate with each other. This signals that \BlonD\ indeed captures additional aspects of translation quality beyond the sentence-level.
Finally, our human evaluation also reveals significantly higher Pearson's $r$ correlation coefficients between \BlonD\ and human assessments compared to other metrics.

\section{\BWB: Bilingual Web Book Dataset} \label{dataset}
To design a metric that is sensitive to document-level phenomena, we first curate a document-level Chinese--English parallel corpus, called \BWB{} (\textbf{B}ilingual \textbf{W}eb \textbf{B}ooks). \BWB{} consists of Chinese web novels across multiple genres (sci-fi, romance, action, fantasy, comedy, \textit{inter alia}) and their corresponding English translations crawled from the Internet.\looseness=-1

\paragraph{Dataset Creation.} The novels are translated by professional native English speakers, and are corrected by editors.
The sentence alignment of the training set is done by Bluealign\footnote{\url{https://github.com/rsennrich/Bleualign}}~\cite{bleualign}. We hired four bilingual graduate students to manually evaluate 163 randomly selected documents from the resulting \BWB\ parallel corpus and observe an alignment accuracy rate of 93.1\%.
We further asked the same batch of annotators to correct such misalignments in both the development and the test set. The details of the corpus creation and quality control are described in \Cref{App:dataset}.\looseness=-1

\paragraph{Statistics.}
\Cref{tab:split} summarizes the statistics of the \BWB{} dataset. It is a much larger dataset, and contains longer documents and richer discourse phenomena compared to all previous document-level datasets~\cite{lison-tiedemann-2016-opensubtitles2016,  koehn-knowles-2017-six, barrault-etal-2019-findings, Koehn2005EuroparlAP,liu-zhang-2020-corpora}. 
To the best of our knowledge, this is the largest Chinese--English document-level translation dataset to date. 
\begin{table}
\centering
\begin{adjustbox}{width=0.46\textwidth}
\begin{tabular}{ccccc}
\toprule[2pt]
\textbf{Statistic} & \textbf{Train} & \textbf{Test} & \textbf{Dev} & \textbf{Total}\\
\midrule[1pt]
\#Docs & 196,304 & 80 & 79 & 196,463 \\
\#Sents & 9,576,566 & 2,632 & 2,618 & 9,581,816\\
\#Words & 325.4M  & 68.0K & 67.4K & 460.8M \\
\bottomrule[2pt]
\end{tabular}
\end{adjustbox}
    \caption{Statistics of the \BWB\ dataset.}
    \label{tab:split}
\end{table}





\paragraph{Dataset Split.} 
We treat chapters in our books as \emph{documents}. The maximum, median, and minimum number of sentences per document are 46, 30 and 18, respectively.
To prevent any train--test leakage, we split the dataset into a training, development and a test set such that chapters from the same book are part of the same split. 
We use 377 books for training, and randomly select 80 and 79 documents from the 3,018 documents in the remaining 6 books as the development and test sets, respectively.\footnote{One document in the development set was dropped due to its poor annotation quality.}

\section{Analyzing Discourse Errors} \label{sec:error_analysis}
\noindent Next, we conduct a human study on the test set of \BWB, in which we identify and categorize the discourse errors made by MT systems that are not captured in sentence-level evaluation. 
This human study is conducted by eight professional translators. The annotators are asked to classify translation errors into \Document-level and \Sentence-level errors (some cases can be both). \Sentence-level errors refer to those errors that render the translations to be inadequate or not fluent as stand-alone sentences, while \Document-level errors reflect a coherence violation across multiple sentences in the document. \Document-level errors are further categorized according to the linguistic phenomena leading to the lower performance in the context-dependent translation.\footnote{The annotation guidelines are described in \Cref{app:error}.}\looseness=-1

\Cref{analysis} shows the result of our error analysis.
A substantial proportion of translations have document-level errors (71.9\%). This supports the claim that \BWB\ contains rich discourse phenomena that common MT systems cannot address. We observe that three categories, i.e., inconsistency (64.4\%), ellipsis (20.3\%) and ambiguity (7.3\%),  account for the vast majority of document-level errors.
Below we discuss these three categories of \Document-level errors and the design intuitions behind \BlonD.\looseness=-1

\begin{table}[t]
\centering \small
\begin{tabular}{p{100pt}cc}
\toprule[2pt]
Error Type & \# & \%\\
\midrule[1pt]
\NoError & 451 & 17.1\% \\
\Sentence & 1351 & 51.3\%\\
\Document & 1893 & 71.9\% \\
\midrule[1pt]
\Inconsistency & 1695 &  64.4\%\\ 
\NamedEntity  & 1139 & 43.3\%\\
\Tense  & 1018 & 38.7\%\\
\Ellipsis  & 534 & 20.3\%\\ 
\Pronoun  & 456 & 17.3\%\\
\Others  & 103 & 4.0\%\\
\Ambiguity  & 193 & 7.3\%\\ 
\bottomrule[2pt]
\end{tabular}
    \caption{The statistics of translation errors in human analysis.}
    \label{analysis}
\end{table}

\paragraph{Inconsistency.}
We consider two kinds of consistency in translation: lexical and grammatical. 
Lexical consistency is defined as a repetitive term that keeps the same translation throughout the whole document~\cite{carpuat-simard-2012-trouble}. 
Inconsistent translation of named entities can significantly impact translation output, although \BLEU\ may not be adversely affected~\citep{agrawal2012using, hermjakob-etal-2008-name}. 
Therefore, in the design of \BlonD, we also focus on the reiteration of named entities (e.g., \Qiao{Qiao} in \Cref{fig:example}).  
On the other hand, typical examples of grammatical consistency are tense and gender consistency. 
Tense consistency refers to the tense being \emph{compatible} with the context, rather than being exactly the same across the whole document.
Tense inconsistency can arise when the source language is an isolating language and does not mark tense explicitly, e.g., in Chinese, and the target language is a synthetic language, e.g., English (\Verb{teal} in \Cref{fig:example}), where tense is marked explicitly.
In the same spirit, the same entity should maintain a consistent grammatical gender.
\footnote{It is worth noting that the metric proposed in this study can be applied to a wider range of language pairs by extending the definition of grammatical consistency.}\looseness=-1

\paragraph{Ellipsis.}
Ellipsis denotes the omission from a clause of one or more words that are nevertheless inferred in the context of the remaining elements \cite{yamamoto,voita-etal-2019-good}. 
Translation errors arise when there are elliptical constructions in the source language while the target language does not allow for the same types of ellipsis. 
For example, the ellipsis of subjects or objects is very common in Chinese while it is ungrammatical in English---especially for pronouns.
In \Cref{fig:example}, \emph{she (Qiao)} is omitted in Chinese. However, it is hard to guess the gender of \emph{Qiao} from this stand-alone sentence: the correct pronoun choice can only be inferred from context (there is a \emph{her} in the previous sentence). 
Another common form of ellipsis is the omission of discourse markers, especially when the source language has more zero connective structures~\cite{po2004chinese} than the target language. In the example, \emph{However} and \emph{So} are ignored in \SRC, which misleads the sentence-level system \MTa\ to ignore the discourse relations between sentences.
\begin{figure}
\centering \small
\begin{tabular}{p{12pt}p{183pt}}
\SRC & \begin{CJK}{UTF8}{gbsn}你在\Verb{看(kan)}什么？《复仇者联盟》。\end{CJK} \\
\REF & What are you \Verb{watching}? The Avengers. \\
\MT & What are you \VerbWrong{looking} at? The Avengers. \\
\end{tabular}
\caption{An example of ambiguity. \begin{CJK}{UTF8}{gbsn}看(kan)\end{CJK} corresponds to \emph{look}, \emph{see}, \emph{watch} and \emph{view}. The correct translation can only be inferred from the next sentence (The Avengers).}
\label{fig:ambiguity}
\vspace{-5pt}
\end{figure}

\paragraph{Ambiguity.}
Translation ambiguity occurs when a word in one language can be translated in more than one way into another language~\cite{tokowicz2010translation}. 
The cross-language ambiguity comes from several sources of within-language ambiguity including lexical ambiguity, polysemy, and near-synonymy. 
A unified feature of these is that ambiguous terms satisfy the form of one-to-many mappings. 
For the example in \Cref{fig:ambiguity}, the word \begin{CJK}{UTF8}{gbsn}看(kan)\end{CJK} can be translated to \emph{look}, \emph{see}, \emph{watch} or \emph{view}. 
Without access to the context, all the lexical choices are sensible. 

\begin{figure*}[t]
\centering \scriptsize
\begin{adjustbox}{width=\textwidth}
\begin{tabular}{llcccc}
\toprule[2pt]
&  & \Entity\ $\entities$ & \Tense\ $\verbs$ & \Pronoun\ $\pronouns$ & \DiscourseMarker\ $\dms$\\
\midrule[1pt]
\SRC &  {\myFontSize a) \begin{CJK*}{UTF8}{gbsn} \Qiao{小乔(Qiao)} \Verb{看着(look)} 相片 \Verb{回忆(recall)} 起了二十年前。\end{CJK*} }& \multirow{4}{*}{ $\llbracket \mathrm{Qiao} \rrbracket$ }& \multirow{4}{*}{\minitab[c]{$[$\texttt{VBD}, \\ \texttt{VBZ}$]$}} &  \multirow{4}{*}{\minitab[c]{$[$\Masculine, \\ \Feminine, \\ \Epicene, \\ \Neuter$]$}} & \multirow{4}{*}{\minitab[c]{$[$\key{contingency}, \\ \key{temporal}, \\ \key{expansion}, \\ \key{comparison} $]$}}\\
&  {\myFontSize b) \begin{CJK*}{UTF8}{gbsn}那个满脸胡须的\Husband{男人(man)} 正\Verb{是(be)}\Qiao{她(she)} 的新婚\Husband{丈夫}。\end{CJK*} } \\ 
&  {\myFontSize c)\begin{CJK*}{UTF8}{gbsn} 那却 \Verb{是(be)} 他们之间初次 \Verb{见面(meet)}。\end{CJK*} }\\
& {\myFontSize d) \begin{CJK*}{UTF8}{gbsn} \Qiao{小乔}\Qiao{(Qiao)}一见到 \Husband{他(he)} 心里就 \Verb{咯噔(jolt)} 了一下，\end{CJK*} }\\
& {\myFontSize \quad \begin{CJK*}{UTF8}{gbsn} 噌的\Verb{站(stand)} 起来。\end{CJK*} }\\
\midrule
\REF &  {\myFontSize a) \Qiao{Qiao} \Verb{looked} at the photo and \Verb{recalled} twenty years ago. } & $[1]$ & $[2,0]$ & $[0,0,0,0]$ & -  \\
 &  {\myFontSize b) This bearded \Husband{man} \Verb{was} \Qiao{her} newlywed \Husband{husband}, } & $[0]$ & $[1,0]$ & $[0,1,0,0]$ &  $[0,0,0,0]$\\
    &  {\myFontSize c) \orangeOmit{yet} this \Verb{was} the first time they \Verb{were meeting} with each other. } & $[0]$ & $[2,0]$ & $[0,0,1,0]$ &  $[0,0,0,1]$ \\
    &  {\myFontSize d) \orangeOmit{So} \Qiao{Qiao}’s heart \Verb{jolted} as soon as \QiaoOmit{she} saw \Husband{him}, and \QiaoOmit{she}}  & $[1]$ & $[2,0]$ & $[1,2,0,0]$ & $[1,0,0,0]$\\
     &  {\myFontSize \quad quickly \Verb{stood} up.} \\
\midrule
\MTa &  {\myFontSize a) \Qiao{Qiao} \Verb{looked} at the photo and \Verb{recalled} twenty years ago. } & $[1]$ & $[2,0]$ & $[0,0,0,0]$ & - \\
    &  {\myFontSize b) This bearded \Husband{man} \VerbWrong{is} \Qiao{her} newlywed \Husband{husband}.} & $[0]$ & $[0,1]$ & $[0,1,0,0]$ &  $[0,0,0,0]$\\
    &  {\myFontSize c) This \VerbWrong{is} the first time they \VerbWrong{meet} with each other. } & $[0]$ & $[0,2]$ & $[0,0,1,0]$ &  $[0,0,0,0]$ \\
    &  {\myFontSize d) \Qiao{\Wrong{Joe}}’s heart \VerbWrong{is} squeaky as soon as \QiaoOmit{\Wrong{he}} saw \Husband{him}, and \QiaoOmit{\Wrong{he}} }  & $[0]$ & $[0,2]$ & $[3,1,0,0]$ & $[0,0,0,0]$\\
     &  {\myFontSize \quad  quickly \VerbWrong{stands} up. } \\
\midrule
\MTb &  {\myFontSize a) \Qiao{Qiao} \Verb{looked} at the photo and \Verb{recalled} the past twenty years ago. } & $[1]$ & $[2,0]$ & $[0,0,0,0]$ & -\\
    &  {\myFontSize b) This \Husband{man} with the beard \Verb{was} \Qiao{her} newly-wed \Husband{husband}. } & $[0]$ & $[1,0]$ & $[0,1,1,0]$ &  $[0,0,0,0]$\\
    &  {\myFontSize c) \orangeOmit{However}, that was the first time they \Verb{met}. } & $[0]$ & $[2,0]$ & $[0,0,1,0]$ &  $[0,0,0,1]$ \\
    &  {\myFontSize d) \orangeOmit{So} as soon as \Qiao{Qiao} saw \Husband{him}, \Qiao{\Omit{her}} heart \Verb{became} squeaky,}  & $[1]$ & $[2,0]$ & $[1,2,0,0]$ & $[1,0,0,0]$\\
    & {\myFontSize \quad and \Qiao{\Omit{she}} swiftly \Verb{stood} up.}  \\
\bottomrule[2pt]
\end{tabular}
\end{adjustbox}
\caption{An example containing inconsistency and ellipsis in \BWB. For inconsistency, the same entities are marked in the same color (\Qiao{Qiao} and \Husband{Husband}), and verbs are marked in \Verb{teal}. For ellipsis, omissions are marked with \Omit{}. \DiscourseMarker\ stands for discourse markers (\orangeOmit{}). The translation mistakes are \underline{underlined}. \MTb\ is intuitively a better system than \MTa\ to human readers.}
\label{fig:example}
\end{figure*}

\section{\BlonD}
The aforementioned document-level phenomena have little impact on the $n$-gram statistics of translations. However, as is shown in \Cref{sec:error_analysis}, they can be key considerations for human readers when evaluating translations at the document level.
Standard automatic metrics ignore the importance of contextual coherence of translations, which implies that the document-level nuances are not being properly modeled~\cite{zhou-etal-2008-diagnostic, xiong}.                                                               
In this section, we describe \BlonD, an automatic metric that explicitly tracks discourse phenomena.\looseness=-1
\subsection{Document-Level Evaluation} \label{sec:Document-Level-Evaluation}

We first give the formulation of measuring discourse phenomena. We define a \defn{document} $\discourse = [S_1, \ldots, S_N]$ as a sequence of $N$ sentences.
We take a \defn{sentence} $\Un$ of length $T$ to be a string of tokens $t_1\cdots t_T$ where each token $t_i$ is taken from the vocabulary $\vocab$.
Let $\spans \left(\Un\right) = \{m_1, m_2, \ldots\}$ be the set of spans in the sentence $\Un$.
Here, a \defn{span} is a subsequence of the tokens in $\Un  = t_1 \cdots t_T$.

Let us assume that we are interested in $K$ \defn{discourse categories}.
Each of these categories capture a discourse phenomenon of interest.
As shown in \cref{sec:error_analysis}, named entity inconsistency, tense inconsistency and pronoun ellipsis make up the majority of discourse errors (67.8\%) on the data analyzed. 
We therefore introduce three types of categories:  \Entity{}, \Tense{} and \Pronoun{}. 
In addition, we introduce \defn{discourse markers} \DiscourseMarker{} as a category, which are the essential contextual links between the various discourse segments (See \Cref{fig:example}).\looseness=-1

For a certain discourse category of interest, $k$, we assume that there are $L_k$ \defn{features}. 
In our case, the features of \Entity{} $\entities$ are a list of named entities in $\discourse$;  the features of \Tense{} are $\verbs = [$\texttt{MD,VBD,VBN,VBP,VBZ,VBG,VB}$],$\footnote{\texttt{MD}: Modal; \texttt{VBD}: Verb (past tense verb); \texttt{VBN}: Verb (past participle); \texttt{VBP}: Verb (non-3rd person singular present); \texttt{VBZ}: Verb (3rd person singular present); \texttt{VBG}: Verb (gerund or present participle); \texttt{VB}: Verb (base form).}
the features of \Pronoun{} are $\pronouns=[$\Masculine, \Feminine, \Neuter, \Epicene $]$;\footnote{\Masculine: he, him, his, himself;  \Feminine: she, her, hers, herself, \Neuter: it, its, itself; \Epicene: they, them, their, theirs, themselves.} the features of \DiscourseMarker{} are $\dms = [\key{contingency}, \key{temporal}, $ $ \key{expansion}, \key{comparison}] $.\footnote{A detailed explanation is provided in \Cref{tab:discourse marker}.} Note that different categories can have different numbers of features and the number of features can be dynamic: $\entities$ depends on $\discourse$ while $\verbs$ and $\pronouns$ are fixed. The intuition behind this is that we want to encourage the system output to keep consistent tense and pronouns as well as the consistent translation for a specific named entity.

Let us now define $\category_{k,l} \left(\Un \right) \subseteq \spans\left(\Un \right) $ as the set of spans in $S$ that share the $l^\text{th}$ feature in the $k^{\text{th}}$ discourse category. To give a concrete example, let us assume that \Tense{} is the $k^{\text{th}}$ category and \texttt{VBD} is the $l^{\text{th}}$ feature in this category. The corresponding $\category_{k,l}(\Un)$ is the set of the spans (in this case, unigrams) tagged with \texttt{VBD} in the sentence $\Un$. In \Cref{fig:example}, all the spans are colored.
We then let $\categories_k(\Un)$ refer to 
a vector of size $L_k$ where each element of that vector is the set $\category_{k,l}(\Un)$.
The sets of spans for \Entity{}, \Tense{}, \Pronoun{} and \DiscourseMarker{} can be produced by an NER model, a POS tagger, a rule-based string match and a discourse marker, respectively.
We also define a weight vector $\bm{w}_k = \left[ w_{k,l} : l \in \{1, \ldots, L_k\} \right]$ for each discourse category $k$, where each entry $w_{k,l}$ corresponds to the weight given to a feature.

We then define the \textbf{discourse representation} of
sentence $\Un$ as the concatenation of all categories:
\begin{equation}
\allcategories(\Un) = \left[ \categories_k \left(\Un \right): k \in \{1,\ldots, K\} \right]
\end{equation}




\paragraph{Similarity.}
Let $\similarity : \categories(\Un^s) \times \categories(\Un^r) \rightarrow \R^K$ represent a similarity vector which measures category-wise similarity between the discourse representations of two sentences $\Un^r$ and $\Un^s$. 
Each entry of the vector $\similarity$ takes non-negative values: The entry being zero if $\Un^s$ and $\Un^r$ have no shared spans with the same discourse category.

The similarity vector $\similarity$ defined here can be implemented in several ways.
A possible implementation of $\similarity$ can be achieved by counting the number of functionally similar spans for each feature and then taking a weighted sum over all features:
\begin{align}
    \similarity(\Un^s, & \Un^r) =   \\
    &    \left[\similarity_k(\Un^s, \Un^r) : k \in  \{1, \ldots, K\}\right]
    \nonumber
\end{align}
where each entry $\similarity_k$ is defined as follows:
\begin{align}
    & \similarity_k(\Un^s, \Un^r) =   \\
    & \ \bm{w}_k \ \odot \  \min\left(\countFun(\categories_k(\Un^s)\right), \countFun(\categories_k(\Un^r)))
    \nonumber
\end{align}
where
\begin{align}
\countFun(&\categories_k(\cdot)) \\ 
&= [ |\category_{k,l}(\cdot)|: l \in \{1, \ldots, L_k \} ] \nonumber
\end{align}
denotes the cardinality of $\category_{k,l}$ applied entry-wise and $\min$ denotes the minimum function applied element-wise. 
Intuitively, $\similarity_k$, measures the number of functionally similar spans shared by $\Un^s$ and $\Un^r$. 
Assume that \Tense{} is the $k^{\text{th}}$ category.\footnote{In \Cref{fig:example}, $\similarity_k(U_b^\mathrm{\MTa}, U_b^\mathrm{\REF}) = 0$ since \MTa\ mistranslated the verbs as present tense due to the exclusion of context. The total similarity 
$\similarity(\discourse^\mathrm{\MTa}, \discourse^\mathrm{\REF})$ is the number of functionally similar spans across all features: (1, 2, 4, 0) for ($\entities$, $\verbs$, $\pronouns$, $\dms$). Here, we assume all category weights are 1.}

It is worth noting that there are many other reasonable ways to operationalize $\similarity$. For \Entity, partial credit could be assigned to two named entities if they have overlapping tokens; for \Tense\ and \Pronoun,  partial credit could be assigned to two similar categories, e.g., \texttt{VBP} and \texttt{VB}; for \DiscourseMarker, partial credit could be assigned according to the sense hierarchy and the confidences in the detected discourse markers. We leave the expansion of the $\similarity$ definition to future work.

\paragraph{A Document-level Similarity Measure.}
Now we turn from measuring the similarity at the sentence level to the document level. 
We first lift $\similarity(\cdot, \cdot)$ to measure the similarity between two documents:
\begin{align}
\similarity\left( \discourse^{s}, \discourse^r \right) &= \sum_{\substack{\Un^s \in \discourse^s, \\ \Un^r \in \discourse^r}} \!\!\! \similarity(\Un^{s}, \Un^{r}) 
\end{align}
where the sum is applied element-wise.

We then define $\similarity(\cdot, \cdot)$ for
a system document $\discourse^s$ and a set of reference documents  $\discourseSet^r = \{\discourse^{r_1}, \discourse^{r_2}, \ldots\}$ by aggregating the $\similarity$ of all sentences in $\discourse^s$ and $\discourseSet^r$:
\begin{equation}
    \similarity\left(\discourse^s, \discourseSet^r \right) = \sum_{\Un^s \in \discourse^s}  \bigoplus_{\Un^r  \in \discourseSet^r}\!\!  \similarity(\Un^s, \Un^r)
\end{equation}

\noindent Here, $\oplus$ is a generic aggregator over multiple references, e.g., $\oplus = \max$, if we take the reference which has the maximum similarity with the system output; or $\oplus = \sum$ if we sum up the similarity scores of all references. Again, $\oplus$ is applied element-wise.\footnote{In \Cref{fig:example}, since we only have one reference,  $\similarity(\discourse^\mathrm{\MTa}, \discourseSet^{r})=\similarity(\discourse^\mathrm{\MTa}, \discourse^\mathrm{\REF})=(1, 2, 4, 0)$.}
\noindent We also reuse the notation $\similarity(\cdot, \cdot)$ for two sets of documents $\discourseSet^{s}$ and $\discourseSet^{r}$:
\begin{align}
\similarity\left( \discourseSet^{s}, \discourseSet^{r} \right) &= \bigoplus_{\discourse^s \in \discourseSet^s} \similarity\left(\discourse^s, \discourseSet^{r} \right)
\end{align}
\noindent Note that the similarity vector can also be computed for the same (set of) documents. For example, if $\similarity$ is implemented as counting the number of functionally similar spans for each feature, then, $\similarity\left( \discourse^{s}, \discourse^{s} \right)$ and $\similarity\left( \discourseSet^{r}, \discourseSet^{r} \right)$ denote the total number of spans of each category in the system output and the reference, respectively.\footnote{In \Cref{fig:example}, $\similarity(\discourse^\mathrm{\MTa}, \discourse^\mathrm{\MTa})=(1, 7, 6, 0)$ and $\similarity\left( \discourseSet^{r}, \discourseSet^{r} \right) = \similarity\left( \discourse^\mathrm{\REF}, \discourse^\mathrm{\REF} \right) = (2, 7, 5, 2)$.}

\paragraph{Scoring.}
We are now ready to define the ``goodness'' of a system output with respect to our discourse phenomenon of interest. 
We compute the precision, recall and F1 for all $K$ discourse categories defined as follows:
\begin{align}
   & \bm{p}(\discourse^s, \discourseSet^r) = \frac{\similarity\left(\discourse^s, \discourseSet^r \right)}{\similarity\left(\discourse^s, \discourse^s \right)},  \\
   & \bm{r}(\discourse^s, \discourseSet^r) = \frac{\similarity\left(\discourse^s, \discourseSet^r \right)}{\similarity\left(\discourseSet^r, \discourseSet^r \right)},  \\
   & \bm{F}(\discourse^s, \discourseSet^r) = 2 \cdot \frac{\bm{p} \odot \bm{r}}{\bm{p}+\bm{r}}.  \label{eq:F1}
\end{align}
\noindent Here, $\bm{p}(\discourse^s, \discourseSet^r)$, $ \bm{r}(\discourse^s, \discourseSet^r)$ and $\bm{F}(\discourse^s, \discourseSet^r)$ are all $K$-dimensional vectors; where, the $k^{\text{th}}$ element of these vectors represents the precision, recall and F-score for the $k^{\text{th}}$ category.
Thus, the addition, multiplication and division operations above are also defined element-wise.\footnote{In \Cref{fig:example}, recall that $\similarity(\discourse^\mathrm{\MTa}, \discourseSet^{r})=(1, 2, 4, 0)$,  $\similarity(\discourse^\mathrm{\MTa}, \discourse^\mathrm{\MTa})=(1, 7, 6, 0)$ and $\similarity\left( \discourseSet^{r}, \discourseSet^{r} \right) = (2, 7, 5, 2)$. Thus, we have $\bm{p}(\discourse^\mathrm{\MTa}, \discourseSet^r) = \left(\frac{1}{1}, \frac{2}{7}, \frac{4}{6}, \frac{0}{0}\right) = \left(1, \frac{2}{7}, \frac{2}{3}, \NA \right)$
where $\NA$ denotes a missing value.
Furthermore, we have
$\bm{r}(\discourse^\mathrm{\MTa}, \discourseSet^r) = \left(\frac{1}{2}, \frac{2}{7}, \frac{4}{5}, \frac{0}{2}\right) = \left(\frac{1}{2}, \frac{2}{7}, \frac{4}{5}, \frac{\delta}{2} \right)$ where $\delta$ denotes a small value (0.0001) for smoothing.
Finally, we have
$\bm{F}(\discourse^\mathrm{\MTa}, \discourseSet^r) = \left(\frac{2}{3}, \frac{2}{7}, \frac{8}{11}, \NA\right).$
}\looseness=-1

\paragraph{\dBlonD.}
Further, we combine the scores of all categories into an overall score with a simple weighted average, named \dBlonD{}. By computing \dBlonD{}, one can distill the document-level translation quality from translation quality at the sentence level.
More formally, we have
\begin{align}
    &\mathrm{\dBlonD.P}(\discourse^s, \discourseSet^r) = \\
    & \qquad { \left(\prod\limits_{k=1}^{K} (\bm{p}_k(\discourse^s, \discourseSet^r)^{a_k} \right) }^{1/\sum\limits_{k=1}^K a_k} \nonumber \\
    &\mathrm{\dBlonD.R}(\discourse^s, \discourseSet^r) = \\
    & \qquad { \left(\prod\limits_{k=1}^{K} (\bm{r}_k(\discourse^s, \discourseSet^r)^{a_k} \right) }^{1/\sum\limits_{k=1}^K a_k} \nonumber
\end{align}
where $a_k $ denotes the importance weight of the $k^{\text{th}}$ category, and $\bm{p}_k$ and $\bm{r}_k$ denote the $k^{\text{th}}$ entry of $\bm{p}$ and $\bm{r}$, respectively.\footnote{\dBlonD\ adopts uniform weights.} 
Therefore, \dBlonD.F1 is defined as follows:
\begin{align}
    &\mathrm{\dBlonD.F1}(\discourse^s, \discourseSet^r) = \\
    & \qquad 2 \cdot \frac{\mathrm{\dBlonD.P} \cdot \mathrm{\dBlonD.R}}{\mathrm{\dBlonD.P} + \mathrm{\dBlonD.R}} \nonumber
\end{align}
\noindent Whenever not otherwise specified, we simply use \dBlonD{} to refer to \dBlonD.F1.\footnote{
\noindent E.g., the \dBlonD{} scores of \MTa{} in \Cref{fig:example} are:
\begin{align}
\mathrm{\dBlonD.P}(\discourse^\mathrm{\MTa}, \discourseSet^r) &= \left(\frac{1}{1}\right)^{\frac{1}{3}}\left(\frac{2}{7}\right)^{\frac{1}{3}}\left(\frac{2}{3}\right)^{\frac{1}{3}}  \nonumber \\
&=.114 \footnote{$\NA$ is ignored.}, \nonumber \\
\mathrm{\dBlonD.R}(\discourse^\mathrm{\MTa}, \discourseSet^r) &= \left(\frac{1}{2}\right)^{\frac{1}{4}}\left(\frac{2}{7}\right)^{\frac{1}{4}}\left(\frac{4}{5}\right)^{\frac{1}{4}} \left(\frac{\delta}{2}\right)^{\frac{1}{4}}  \nonumber \\
&=.057, \nonumber \\
\mathrm{\dBlonD.F1}(\discourse^\mathrm{\MTa}, \discourseSet^r) &= \frac{2\cdot .057 \cdot .114}{.057 +.114}  \nonumber \\
&= .076. \nonumber
\end{align}}

\begin{table}[t]
    \centering \small
    \begin{tabular}{c|c|ccc|c}
    \toprule[2pt]
    ~ & \BLEU & \multicolumn{3}{c|}{\BlonD} & \dBlonD \\
    ~& P & P & R & F1 & F1\\
    \midrule[1pt]
    \MTa &41.5 & 10.5 & 51.3 & 17.4 & 7.6 \\
    \MTb &35.9 & 60.6 & 58.9 & 59.8 & 97.7 \\
    \bottomrule[2pt]
    \end{tabular}
    \caption{The \BLEU\ and \BlonD\ scores of the two system outputs in \Cref{fig:example}. P, R and F1 represent precision, recall and F1, respectively. }
    \label{tab:example:blond_scores}
    \vspace{-5pt}
\end{table}

\subsection{\BlonD: Combining \dBlonD with $n$-grams}
However, focusing on discourse phenomena solely is not enough to provide comprehensive MT evaluation that correlates strongly with human judgments. Consider the following example:
\ex. \label{ex:n_gram_example}
\a.[\REF] Qiao lifted her heavy eyelids.
\b.[\MT] Qiao scrunched her brows together. 

The output of \MT\ is far from ``good'' in terms of adequacy, whereas  $\mathrm{\dBlonD}(\mathrm{\MT})=1$, since \MT\ translates both named entities and tenses correctly.
Thus, in order to account for sentence-level adequacy of our final metric \BlonD{},
we augment the set of categories and features to include each $n$-gram (for a value of n) as a category and each span of $n$-tokens as a feature for the $n$-gram category.
Formally, we have
\begin{align}
&\allcategories'(\Un) = \\
&\quad\quad \left[ \categories_k \left(\Un \right) : k \in \{1, \ldots, K+N\} \right] \nonumber
\end{align}
where we define
\begin{equation}
\categories_{K+n} = \left\{n\text{-gram}: n \in \{1, \ldots, N\} \right\}
\end{equation}

\noindent The calculation of \BlonD.P, \BlonD.R and \BlonD.F1 is then done exactly in the same manner as \dBlonD{}. Whenever not specified, we simply use \BlonD{} to refer to \BlonD.F1.

\BlonD{} covers both discourse coherence features and sentence-level adequacy, thus providing a comprehensive measurement of translation quality.
\Cref{tab:example:blond_scores} compares \BlonD\ with \BLEU\ using the two MT outputs found in \Cref{fig:example}. 
It is striking that \BLEU\ rates \MTa\ higher than \MTb\, given that \MTb\ is clearly better than \MTa\ to human readers.
In sharp contrast, their \BlonD\ scores reflect the correct ranking in translation quality.
\newcommand{\setfigheight}{55pt}
\newcommand{\imgtrimtopsize}{1cm}
\newcommand{\imgtrimleftsize}{1cm}
\newcommand{\imgtrimrightsize}{1cm}
\newcommand{\imgtrimbottomsize}{2cm}
\newcommand{\setfigwidth}{\textwidth}

\begin{figure*}[t]
    \centering
  \includegraphics[width=\setfigwidth,trim={0.5cm 10cm 1cm 8cm} ,clip]{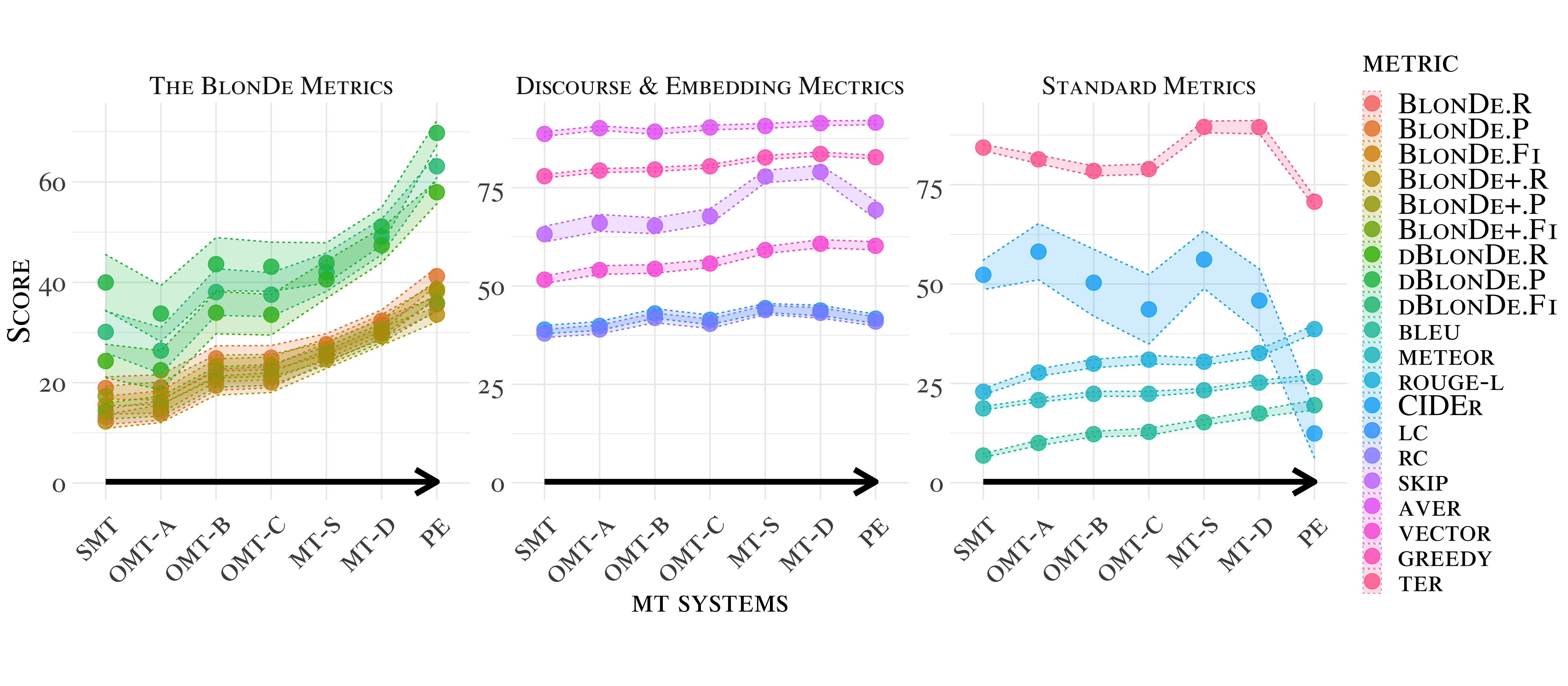}
    \caption{The mean scores of different system outputs given by different metrics on the \BWB\ test set. Shaded region represents 95\% confidence interval.\looseness=-1}
    \label{fig:scores}
\end{figure*}

\subsection{\BlonDp: Combining \BlonD with Human Annotations}
\BlonD\ is easy to generalize---for instance, it would be easy to incorporate human annotations, e.g., one could annotate spans related to discourse errors and treat them as categories. 
The automatically inferred categories and human annotated categories are then combined by adopting the same weighted averaging approach, which we call \BlonDp.
We hired the same translators who analyzed discourse errors in \Cref{sec:error_analysis} to annotate ambiguous and omitted word/phrases on the test set of \BWB.\footnote{We also make this annotated test set publicly available as a testbed for evaluating the ability of MT systems to disambiguate word senses and to predict coherent pronouns or discourse markers in the case of omission.}

\section{Experiments}
In this section, we examine the effectiveness of \BlonD\ at the document-level MT evaluation through experiments. We answer the following question: Do differences in \BlonD\ reliably reflect differences in the document-level translation quality of different MT systems?
To answer this question, we run several MT baselines and compare their \BlonD\ scores to eleven other metrics:
\paragraph{Standard Sentence-level Metrics.} 
\BLEU\ \cite{BLEU}, \METEOR\ \cite{Meteor}, \TER\ \cite{TER}, \ROUGEL\ \cite{rouge}, \CIDEr\ \cite{CIDEr}.
\paragraph{Document-level Metrics.}
\LC\ and \RC\ \cite{wong-kit-2012-extending}---these are ratios between the number of lexical cohesion devices (repetition and collocation) and repeated content words over the total number of content words in a target document, which are direct measurements of lexical cohesion. 

\paragraph{Embedding-based Metrics.} 
We consider four embedding-based metrics in this work: SkipThought cosine similarity \cite[\Skip;][]{SkipThought}, embedding average cosine similarity \cite[\Aver ;][]{nlgeval}, Vector extrema cosine similarity \citep[\Vector;][]{Vector}, Greedy Match \citep[\Greedy  ;][]{Greedy}.

\subsection{MT Systems}
We test \BlonD{} on the following system outputs: an \SMT\ system~\cite{SMT}, three well-known commercial NMT systems (\OMTa, \OMTb, \OMTc), a sentence-level transformer-based system (\MTS) and a document-level system (\MTD) trained on \BWB. \MTD{}~\cite{ctx} trains sentence-level model parameters and then estimates document-level model parameters while keeping the sentence-level Transformer model parameters fixed. We adopt Transformer Big \cite{NIPS2017_7181} for both \MTS\ and \MTD.
The final ``system'' is a human post-editing (\PE) on \OMTc{} provided by professional translators, so it is supposed to be the strongest baseline.\footnote{We trained models by fairseq~\cite{fairseq}. Model parameters and the post-editing details are in Appendix \ref{app:model_parameters} and \ref{app:pe}, respectively.}


\subsection{The \BlonD\ Evaluation}
Firstly, we leverage the test set of \BWB{} and evaluate the above-mentioned systems by \BlonD\ and other metrics. 
\Cref{fig:scores} presents the means of all metrics along with the 95\% confidence interval estimated from bootstrap resampling.
We observe that the \BlonD{} scores demonstrate an exponentially increasing trend from sentence-level towards document-level and human post-editing, while the trends of standard metrics are mostly linear. Specifically, the difference between the \BlonD{} scores of \MTS{} and \MTD{} (denoted as \diff{\MTS}{\MTD}) is significantly higher than the difference between the \diff{\MTS}{\MTD} in their \BLEU{} scores. An even larger $\Delta$ between \MTD{} and \PE{} in their \BlonD{} scores is observed, indicating \MTD{} is still far away from achieving human parity. 
Note that the trend of \dBlonD{} scores is even more exponential, which indicates that \dBlonD{} indeed distills document-level translation quality.


\begin{table*}[t]
\centering \small
\begin{tabular}{c|c|ccc|ccc|ccc|cccc}
\toprule[2pt]
&\multirow{2}{*}{\BLEU}  &  \multicolumn{3}{c|}{\BlonD} & \multicolumn{3}{c|}{\BlonDp} & \multicolumn{3}{c|}{\dBlonD} & \multicolumn{4}{c}{Categories}\\
~ & & R & P & F1 & R & P & F1 & R & P & F1 & $\entities$ & $\verbs$ & $\pronouns$ & $\dms$\\
\midrule[1pt]
\SMT $\rightarrow$ \MTS & 25.8     & 13.5     & 7.42     & 10.9      & \textbf{14.5}      & 8.51      & 12.0         & 8.02      & \gray{1.32}      & 5.10  & -2.12                           & 23.6                            & 11.4       & 13.6 \\
\MTS $\rightarrow$ \MTD & 8.97      & 6.32     & 5.45     & 5.92      & \textbf{6.58}      & 5.61      & 6.13       & 4.85      & 4.57      & 4.79 & 4.93                            & \gray{1.88}                            & 7.43       & \gray{1.62}\\

\MTD $\rightarrow$ \PE &	2.6      & 4.51     & 7.77     & 6.06     & 4.20       & 7.27      & 5.66       & 6.44      & \textbf{11.1}      & 8.58 & 12.9                            & 2.44                            & 2.76      & 5.35 \\
\bottomrule[1pt]
\end{tabular}
\begin{tabular}{c|cccc|cc|cccc}
\toprule[1pt]
 &\multicolumn{4}{c|}{Other Standard Metrics} & \multicolumn{2}{c|}{Discourse Cohesion} & \multicolumn{4}{c}{Embedding-based Metrics}\\
~ & \METEOR & \ROUGEL & \TER & \CIDEr & \LC & \RC & \Skip & \Aver & \Vector & \Greedy \\
\midrule[1pt]
\SMT $\rightarrow$ \MTS & 25.3      & 19.8      & -8.28     & \gray{.853}     & 11.7      & 12.9      & 12.2      & 9.50       & 18.0        & 22.3      \\

\MTS $\rightarrow$ \MTD & 13.4     & 11.8     & \gray{.148}    & -3.03    & \gray{-1.23}    & \gray{-1.45}    & \gray{1.62}     & 3.13     & 5.05     & 5.83 \\
\MTD $\rightarrow$ \PE &3.58     & 9.65     & 19.9     & -6.67    & -4.23    & -4.44    & -6.23    & \gray{.628}    & \gray{-1.03}    & -3.15\\
\bottomrule[2pt]
\end{tabular}
    \caption{The paired $t$-statistics of different MT systems. The cells with $p$-value $>.05$ are marked in gray. While \BLEU{} distinguishes \SMT{} and the sentence-level \MTS{} significantly, it fails to possess the same discriminative power towards document-level and human translations. \BlonD{} maintains similar discriminative power across the three $t$-tests. }
    
    \label{tab: t-statistics}
\end{table*}
\begin{figure}[t]
    \centering
  \includegraphics[width=0.49\textwidth,trim={1cm 8cm 1cm 6cm} ,clip]{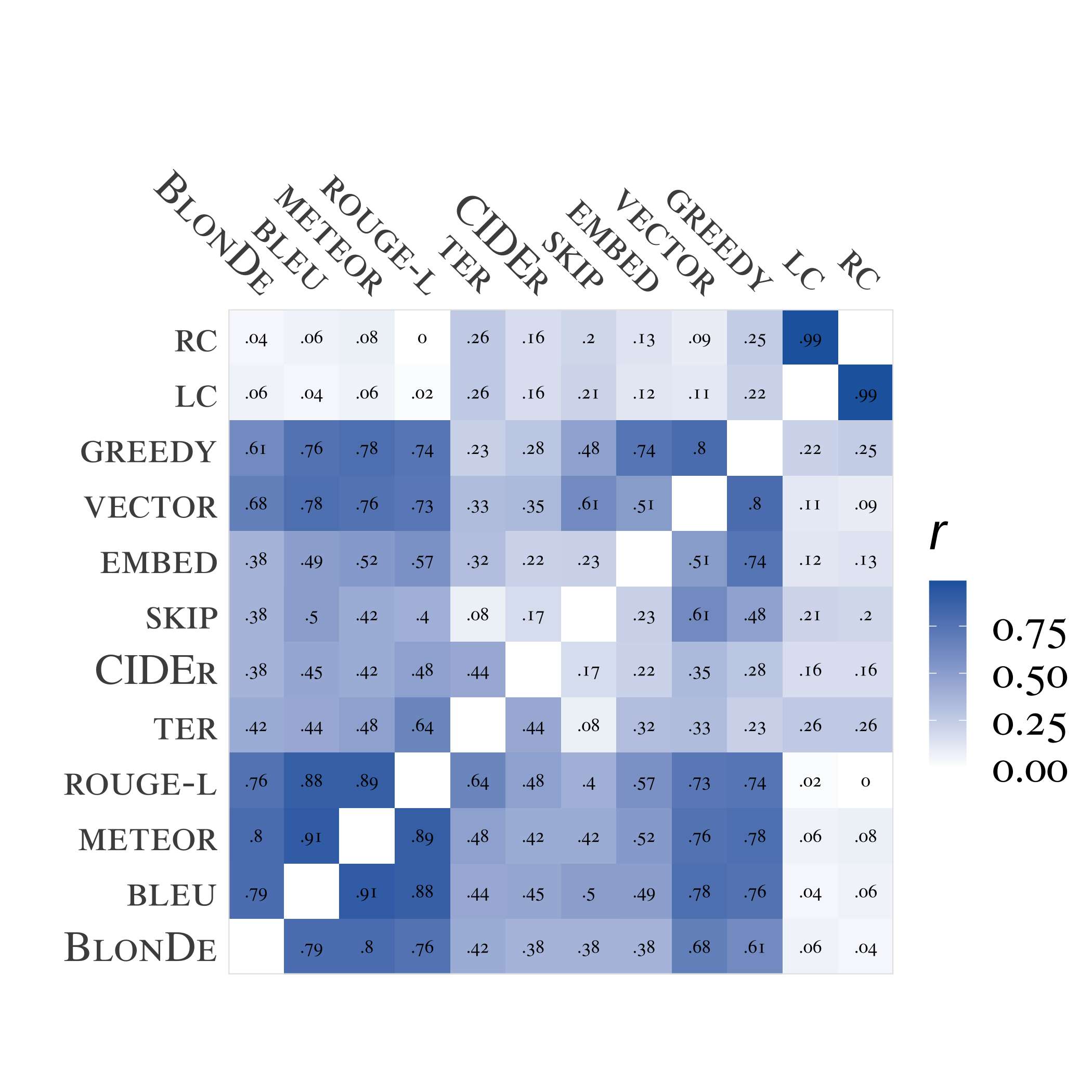}
    \caption{Absolute Pearson correlation pairs of automatic metrics. Computed over the scores of individual documents in \BWB{} test set.\looseness=-1
    }
    \label{fig:correlation}
\end{figure}

The $t$-statistics of the paired sample $t$-tests of individual documents are given in \Cref{tab: t-statistics}.
Unlike \BLEU{}, \METEOR{} and other metrics, which either fails to distinguish human and machine translation or has lower discriminative power compared to distinguishing different machine translations, the \BlonD{} family maintain similar discriminative power across the pair-wise comparisons. 
Interestingly but not surprisingly, the non-reference-based \LC{} and \RC{} fail to distinguish both (\MTS, \MTD) and (\MTD, \PE), since sentence-level MT is by nature more repetitive than human translation and thus it is hard to distinguish accidental repetition from document-level cohesion.

In addition, the $t$-statistics of \dBlonD{} categories provide rich diagnostic information. As can be seen, although transformer-based NMT models have substantially higher \BLEU{} scores than \SMT{} systems, 
\MTS{} is not statistically superior to \SMT{} in terms of named entity translation. However, 
human post-editing scores significantly better on entity translation---meaning that named entity translation accounts for a substantial part of quality differences between machine and human.
In terms of \Tense{} and and \DiscourseMarker{} translation, \MTD{} does not significantly out-perform \MTS{}, which could be taken into consideration in future document-level MT model designs.

\begin{table}[t]
\centering \small
\begin{tabular}{c|cc|cc}
\toprule[2pt]
\multirow{2}{*}{Unit}  & \multicolumn{2}{c|}{\Sentence} & \multicolumn{2}{c}{\Document} \\
& \textsc{ade} & \textsc{flu} & \textsc{ade} & \textsc{flu}\\
\midrule[1pt]
\BlonD.R   & \textbf{.363}  & .327           & .436$^\dag$        & .371$^\dag$       \\
\BlonD.P   & .331           & .296           & .383$^\dag$        & .344$^\dag$       \\
\BlonD.F1  & .35            & .314           & .417$^\dag$        & .358 $^\dag$      \\
\BlonDp.R  & .364           & \textbf{.329}  & \textbf{.44}$^\dag$         & \textbf{.373}$^\dag$       \\
\BlonDp.P  & .334           & .3            & .39$^\dag$         & .349$^\dag$       \\
\BlonDp.F1  & .351           & .318           & .422$^\dag$        & .362$^\dag$       \\
\midrule[1pt]
\BLEU      & .325           & .308           & .343        & .266       \\
\METEOR    & .338           & .31           & .339        & .278       \\
\ROUGEL    & .275           & .262           & .29         & .211       \\
\TER       & .063           & .027           & .044        & .092       \\
\CIDEr     & .139           & .116           & .114        & .087       \\
\Skip      & .213           & .174           & .163        & .171       \\
\Aver      & .163           & .163           & .16         & .111       \\
\Vector    & .25            & .243           & .248        & .218       \\
\Greedy    & .323           & .3            & .307        & .265       \\
\midrule[1pt]
\LC        & .086           & .061           & .153        & .116       \\
\RC        & .096           & .07           & .169        & .13        \\
\bottomrule[2pt]
\end{tabular}
    \caption{Absolute Pearson correlation with human judgments on \BWB{}. The highest correlations are in bold. Correlation of metrics not significantly outperformed by any other metrics are highlighted with $\dag$. The \BlonD{} family are not tested against each other.}
    \label{tab:human_pearson_correlation}
\end{table}

We also show the pairwise Pearson correlations between different metrics in \Cref{fig:correlation}. It illustrates the homogeneity/heterogeneity of different metrics. 
We report the absolute value of the correlation for \TER.
We see that while sentence-level metrics (\BLEU, \METEOR\ and \ROUGEL) have strong correlations with each other, \BlonD{} correlates less well with those metrics.

\subsection{Human Evaluation}
We then evaluate \BlonD{} along with other metrics in terms of their Pearson correlation with human assessment. Our human assessment is provided by four professional Chinese to English translators and four native English revisers. Two experimental units (\Sentence{} vs \Document{}) are assessed independently in terms of \fluency{} and \adequacy{}, respectively. 
In the \Sentence-level evaluation, we show the raters isolated sentences, while in the \Document-level evaluation, we show them entire documents and we only ask raters to evaluate the overall quality of sequential blocks of sentences (5 sentences per block) as used in the Relative Ranking (RR) evaluation~\cite{bojar-etal-2016-results}.
We use the Williams significance test~\citep{williams1959regression, graham-baldwin-2014-testing} following the practice adopted by WMT~\cite{WMT2020-results} to identify correlation differences are statistically significant.
The detailed protocol is presented in \cref{app:human_evaluation}.\looseness=-1


The results are shown in \Cref{tab:human_pearson_correlation}. \BlonD\ obtains the highest correlation with human assessment at both the sentence level and the document level. 
However, \BlonD{} correlates remarkably better with human assessment when context is taken into account, and it only significantly outperforms all other metrics at document level. 

It is worth noting that \BlonD\ also correlates well with \fluency{} assessment, even though it is, in essence, still a reference-based metric. One possible explanation for this unexpected positive result is that it tracks span categories that directly relate to cohesion and coherence.
Another important observation is that the recall-based \BlonD{} variants generally correlate better with human assessment, yet appears to be less selective compared to the precision-based variants (see \MTD $\rightarrow$ \PE{} in \Cref{tab: t-statistics}). This provides support for adopting the F1 in order to get the best of both worlds.

\section{Related Work}

There have been a few studies on automatic evaluation metrics for specific discourse phenomena. 

\paragraph{Pronoun Translation.} \citet{hardmeier2010modelling} measured the precision and recall of pronouns directly and \citet{miculicich} proposed to estimate the accuracy of pronoun translation (APT) by aligning source and target texts. 
However, as shown in \citet{guillou-hardmeier-2018-automatic}, APT does not take the antecedents of an anaphoric pronoun into account. They cannot handle the mismatches in the numbers of pronouns either.
\citet{jwalapuram2019evaluating} also proposed a specialized measure for pronoun evaluation which involves training.
In comparison, \BlonD{} does not rely on any alignment or training.\looseness=-1

\paragraph{Lexical Cohesion.} \citet{wong-kit-2012-extending} proposed \textsc{lc} and \textsc{rc}. \citet{gong-etal-2015-document} described a cohesion function to measure text cohesion via lexical chain and a gist consistency score based on topic model. However, they fail to distinguish accidental repetition from document-level cohesion. 

\paragraph{Discourse Relations.} \citet{hajlaoui2013assessing} proposed to assess the accuracy of connective translation (ACT).
However, such an assessment requires a bilingual dictionary of all possible DM translations, whereas \BlonD{} only requires a list of monolingual DMs. 
\citet{guzman-etal-2014-using} and \citet{joty-etal-2014-discotk} compute a metric based on the similarity between the discourse trees of reference and system output. 
Those discourse-representation-based metrics are indirect, and rely on discourse parsing tools, which are much more inaccurate than syntactic and semantic parsing tools used in \BlonD.
Unlike previously proposed metrics, \BlonD\ does not only focus on one specific discourse phenomenon, and thus has significantly higher Pearson correlation coefficients with human assessments.

\section{Conclusion}
In this paper, we build a large-scale parallel dataset for document-level translation, \BWB. We analyze it for common document-level translation errors in practice and
propose \BlonD, an interpretable automatic metric for document-level MT evaluation.
We further improve \BlonD{} by diagnosing and distilling discourse-related errors in MT outputs and human-annotations to obtain two improved metrics \dBlonD{} and \BlonDp. These metrics were shown to have better selectivity than various sentence-level metrics and correlate better with human judgments.

\section*{Ethical Considerations}
The annotators were paid a fair wage and the annotation process did not solicit any sensitive information from the annotators. 
Finally, while our approach is not tuned for any specific real-world application, the approach could be used in
sensitive contexts such as legal or health-care settings, and any work must use our approach undertake extensive quality-assurance and robustness testing before using it in their setting.

\paragraph{Replicability.} 
As part of our contributions, we
will release the annotated \BWB{} test set, and release the crawling script of the training set under Fair Use rules. The \BlonD{} package is also publicly available at \url{https://github.com/EleanorJiang/BlonDe}.

\section*{Acknowledgements}
Mrinmaya Sachan acknowledges support from an ETH Z\"urich Research grant (ETH-19 21-1) and a grant from the Swiss National Science Foundation
(project \#201009) for partially supporting this work.\looseness=-1

\bibliography{blond}
\bibliographystyle{acl_natbib.bst}

\clearpage
\begin{appendices}

\begin{table*}[t]
\centering \small
\begin{tabular}{p{2cm} p{5cm} p{7cm}}
\toprule[2pt]
\textbf{CATEGORIES} & \textbf{DESCRIPTION}  & \textbf{MARKERS}\\
\midrule[1pt]
\key{contingency} & only consider ``cause" &  [``but", ``while", ``however", ``although", ``though", ``still", ``yet", ``whereas", ``on the other hand", ``in contrast", ``by contrast", ``by comparison", ``conversely"]\\
\midrule[1pt]
\key{comparison}& combine ``concession'' and ``contrast'' & [``if'', ``because'', ``so'', ``since'', ``thus'', ``hence'', ``as a result'', ``therefore'', ``thereby'', ``accordingly'', ``consequently'', ``in consequence'', ``for this reason'']\\
\midrule[1pt]
\key{expansion}&  only consider ``conjunction''  & [``also'', ``in addition'', ``moreover'', ``additionally'', ``besides'', ``else,'', ``plus''] \\
\midrule[1pt]
\multirow{2}{*}{\key{temporal}}& ``synchronous''  &[``meantime'', ``meanwhile'', ``simultaneously'']\\
 & ''asynchronous'' &  [``when'', ``after'', ``then'', ``before'',
                     ``until'', ``later'', ``once'', ``afterward'', ``next''] \\
\bottomrule[2pt]
\end{tabular}
    \caption{Explanations of the discourse marker types (discourse relations) in \DiscourseMarker.}
    \label{tab:discourse marker}
    \vspace{0em}
\end{table*}

\begin{table*}[htpb]
\centering
\begin{tabular}{cccccc}
\toprule[2pt]
\textbf{Dataset} & \textbf{Domain} & \textbf{\#Docs} & \textbf{\#Sents} \\
\midrule[1pt]
WMT~\cite{barrault-etal-2019-findings} & News & 68.4k & 3.63M  \\
OpenSubtitles~\cite{opensubtitles2018} & Subtitles & 29.1k & 31.2k \\
TED~\cite{iwslt2020} & Talks & 1K & 219M  \\
\midrule[1pt]
\BWB\ & Books  & 196k & 9M  \\
\bottomrule[2pt]
\end{tabular}
    \caption{Comparison of different document-level datasets.}
    \label{compare}
\end{table*}


\section{Dataset Creation} \label{App:dataset}
\paragraph{The Background of Translators.}
The original Chinese books are translated by professional native English speakers, and are corrected by editors. 

\paragraph{Data Collection.}
This process is implemented by a python web crawler, and certain data cleaning is also done in the process. We crawl the books chapter by chapter, and convert the text to UTF-8. After deduplication, we remove the chapters with less than 5 sentences. We further remove the titles of each chapter, because most of them are neither translated properly nor in the document-level.

\paragraph{Alignment and  Quality Control.}
After collecting the web books, we align the bilingual books chapter by chapter according to the indices, while removing those chapters without parallel data. 
Then, we use Bluealign, which is an MT-based sentence alignment tool, to align the chapters into parallel sentences, while retaining the document-level information.
We further deduplicate the parallel corpus and filter the pairs with a sequence ratio of 3.0. 
The scale of the final corpus is 384 books with 9,581,816 sentence pairs (a total of 460 million words). 
To estimate the accuracy of this process, we hired 4 bilingual graduate students to manually evaluate 163 randomly selected documents from the resulting \BWB\ parallel corpus. These students are native Chinese speakers who are proficient in English. More specifically, they were asked to distinguish whether a document is well aligned at the sentence level by counting the number of misalignment. For example, if Line 39 in English actually corresponds to Line 39 and Line 40 in Chinese, but the tool made a mistake that it combines the two sentences, it is identified as a misalignment.
We observe an alignment accuracy rate of 93.1\%.

We further asked the same batch of annotators to correct such misalignments in both the development and the test set. 
The annotation result shows that 7.3\% lines are corrected.

\section{Error Analysis and \BlonDp\ Annotation}\label{app:error}
Error analysis and \BlonDp\ annotation are conducted together. This task is conducted by eight professional Chinese-English translators who are native in Chinese and fluent in English.

The guideline is as follows:
\begin{itemize}
    \item First, identify cases which have translation errors. The annotators are instructed to mark examples as “translations with no error” only if it satisfies the criteria of both adequacy and fluency as well as satisfies the criterion that it is coherent in the context.
    \item Second, identify whether the translation contains document-level error or sentence-level error (or both). The annotators are instructed to mark examples as “cases with sentence-level errors” when they are not adequate or fluent as stand-alone sentences; while “document-level errors” mean those errors that cause the example violating the global criterion of coherence.
    \item Third, categorize the examples with document-level errors according to the linguistic phenomena that lead to errors in MT outputs when considering context. 
\end{itemize}

We first conduct a test annotation and observe that the annotators categorize document-level errors into mainly into 3 categories, namely inconsistency, ellipsis, and ambiguity. According to this observation, we instruct annotators to mark document-level errors as \textit{inconsistency}, \textit{ellipsis}, and \textit{ambiguity}, or \textit{other document-level error} during the annotation process for the entire test set. 

In the formal annotation process, we also added the requirement to annotate \BlonDp\ spans. The detailed requirement is as follows:
\begin{itemize}
    \item Third, categorize the examples with document-level into 4 categories: inconsistency, ellipsis, and ambiguity, or other document-level error which cannot be categorized. 
    \item Fourth, if the example is categorized as \textit{ambiguity}, mark the specific word/phrase in the reference (English) that cause ambiguity and give the correct word/phrase.
    \item Fifth, if the example is categorized as \textit{ellipsis} and it is \emph{not} related to pronouns or discourse markers, mark the omitted word/phrase in the reference (English).
\end{itemize}






\section{Human Post-Editing} \label{app:pe}
This task is conducted by the same eight professional Chinese-English translators who carry out the annotation in \Cref{app:error}. We asked them to follow guidelines for achieving “good enough” quality at the sentence-level (comprehensible, accurate but as not being stylistically compelling) but especially pay attention to document-level errors and correct them.

\section{The Human Evaluation Protocol} \label{app:human_evaluation}
The human evaluation is conducted on outputs of four systems (\OMTb, \MTS, \CTX, \PE) and human translation. We follow the protocol proposed by \citep{laubli-etal-2018-machine, laubli2020set}. 
We conduct the evaluation experiment with a $2 \times 2$ mixed factorial design, carrying both \Document-level and \Sentence-level evaluation in terms of \adequacy\ and \fluency. 
In the \Sentence-level evaluation, we show raters isolated sentences in random order;
while in the \Document-level evaluation, entire documents are presented and we only ask raters to
evaluate a sequence of 5 sequential sentences at a time in order. 

To avoid reference bias, the \adequacy\ evaluation is only based on source texts, while no source texts nor references are presented in the \fluency\ evaluation.
We adopt Relative Ranking (RR): raters are presented with outputs from the aforementioned five systems, which they are asked to evaluate relative to each other, e.g., to determine
system A is better than system B (with ties allowed).

We use source sentences and documents from the \BWB\ test set, but blind their origins by randomizing both the order in which the system outputs are presented. Note that in the \Document-level evaluation, the same ordering of systems is used within a document. The order of experimental items is also randomised. 
Sentences are randomly drawn from these documents, regardless of their position.

We also use spam items for quality control \cite{kittur2008crowdsourcing}: In a small fraction of items, we render one of the five options nonsensical by randomly shuﬄing the order of all translated words, except for 10\% at the beginning and end. If a rater marks a spam item as better than or equal to an actual translation, this is a strong indication that they did not read both options carefully. At the \Document-level, we render one of the five options nonsensical by randomly shuﬄing the order of all translated sentences, except for the first and the last sentence. 

We recruit four professional Chinese to English translators and four native English revisers for the adequacy and fluency conditions respectively.
Note that the eight translators are different from those professional translators who carry out the human translation \PE. We deliberately invite another group of specialists for human evaluation to avoid making unreasonable judgments biased towards \PE. 
In each condition, each raters evaluate 162 documents (plus 18 spam items) and 162 sentences (plus 18 spam items).
We use two non-overlapping sets of documents and two non-overlapping sets of sentences, and each is evaluated by two raters.
Specifically, we refer the first half of the test set as \testset{1} and the second half as \testset{2}. Note that \testset{1} and  \testset{2} are chosen from different books. Each rater evaluates both sentences and documents, but never the same text in both conditions so as to avoid repetition priming \cite{gonzalez2011cognitive}: \rater{1} and \rater{2} conduct the \Document-level \adequacy\ evaluation on 180 documents sampled from \testset{1} and the \Sentence-level \adequacy\ evaluation for \testset{2}; \rater{3} and \rater{4} conduct the \Sentence-level \fluency\ evaluation on 180 documents sampled from \testset{1} and the \Document-level \fluency\ evaluation for \testset{2}; \rater{5} and \rater{6} conduct the \Document-level \fluency\ evaluation on 180 documents sampled from \testset{1} and the \Sentence-level \fluency\ evaluation for \testset{2}; \rater{7} and \rater{8} conduct the \Sentence-level \fluency\ evaluation on 180 documents sampled from \testset{1} and the \Document-level \fluency\ evaluation for \testset{2}.

\section{Statistical Analysis of Human Evaluation}
We calculate Cohen's kappa coefficient:
\begin{equation}
    \kappa = \frac{P(A)-P(E)}{1-P(E)}
\end{equation}
where $P(A)$ is the proportion of times that two raters agree, and $P(E)$ is the likelihood of agreement by chance. We report pairwise inter-rater agreement in \Cref{kappa}.
\begin{table}[]
\centering
\begin{tabular}{ccc}
\toprule[2pt]
              & \Sentence & \Document  \\ 
\midrule[1pt]
\rater{1}-\rater{2} & .171 & .169 \\
\rater{3}-\rater{4} & .294 & .346 \\
\rater{5}-\rater{6} & .323 & .402 \\
\rater{7}-\rater{8} & .378 & .342 \\ 
\bottomrule[2pt]
\end{tabular}
\caption{Inter-rater agreements measure by Cohen's $\kappa$, where \rater{1-4} are professional translators whose native language is Chinese, \rater{5-8} are native English revisers.}
\label{kappa}
\end{table}

\section{Experiment Settings}
\subsection{\BlonD}
We use the named entity recognition module and the POS tagger of spaCy~\cite{spacy2} to implement the categorizing function $\categorizing$ for \Entity\ and \Tense, respectively. We use the script provided by \citet{sileo-etal-2019-mining} as the discourse marker minor.

\subsection{Model Hyperparameters} \label{app:model_parameters}
We follow the setup of Transformer big model for \BWB{} experiments. More precisely, the parameters in the big encoders and decoders are $N=12$ , the number of heads per layer is $h = 16$, the dimensionality of input and output is $d_{model} = 1024$, and the inner-layer of a feed-forward networks has dimensionality $d_{ff} = 4096$. The dropout rate is fixed as 0.3. We adopt Adam optimizer with $\beta_1 = 0.9, \beta_2 = 0.98, \epsilon = 10^{-9}$, and set learning rate $0.1$ of the same learning rate schedule as Transformer. We set the batch size as 6,000 and the update frequency as 16 for updating parameters to imitate 128 GPUs on a machine with 8 V100 GPU. The datasets are encoded by BPE with 60K merge operations.


\end{appendices}

\end{document}